\documentclass[letterpaper, 10 pt, journal, twoside]{IEEEtran}

\usepackage{cite}
\usepackage{amsmath,amssymb,amsfonts}
\usepackage{algorithmicx}
\usepackage{algorithm}
\usepackage{algpseudocode}
\usepackage{mathtools}
\usepackage{algpseudocode}
\usepackage{graphicx}
\usepackage{textcomp}
\usepackage{xcolor}
\usepackage{subfigure}
\usepackage{multirow}
\usepackage{colortbl}
\usepackage{bm}
\usepackage{amsthm} 
\usepackage{algorithm}
\usepackage{threeparttable}
\usepackage{booktabs}
\usepackage{balance}
\usepackage{float}
\usepackage{array}
\def\BibTeX{{\rm B\kern-.05em{\sc i\kern-.025em b}\kern-.08em
    T\kern-.1667em\lower.7ex\hbox{E}\kern-.125emX}}

\definecolor{Gray}{gray}{0.9}

\begin{document}

\title{Disturbance-Aware Adaptive Compensation in Hybrid Force-Position Locomotion Policy for Legged Robots}

\author{Yang Zhang$^{1}$, Buqing Nie$^{2}$, Zhanxiang Cao$^{2}$, Yangqing Fu$^{2}$, and Yue Gao$^{3\dag}$
\thanks{This work was supported by the National Natural Science Foundation of China (Grant No. 62373242 and No. 92248303), the Shanghai Municipal Science and Technology Major Project (Grant No. 2021SHZDZX0102), and the Fundamental Research Funds for the Central Universities.}
\thanks{$^{1}$Yang Zhang is with Department of Automation, Shanghai Jiao Tong University, Shanghai, P.R. China. Email: zhangyang-sjtu-2022@sjtu.edu.cn}
\thanks{$^{2}$Buqing Nie, Zhanxiang Cao, and Yangqing Fu are with Department of Computer Science, Shanghai Jiao Tong University, Shanghai, P.R. China. Email: niebuqing@sjtu.edu.cn, caozx1110@sjtu.edu.cn, and frank79110@sjtu.edu.cn}
\thanks{$^3$Yue Gao is with MoE Key Lab of Artificial Intelligence and AI Institute, Shanghai Jiao Tong University, Shanghai, P.R. China. Email: yuegao@sjtu.edu.cn}
\thanks{$\dag$ Corresponding author.}
}

\maketitle
\pagestyle{empty}  
\thispagestyle{empty} 

\begin{abstract}
Reinforcement Learning (RL)-based methods have significantly improved the locomotion performance of legged robots.
However, these motion policies face significant challenges when deployed in the real world.
Robots operating in uncertain environments struggle to adapt to payload variations and external disturbances, resulting in severe degradation of motion performance.
In this work, we propose a novel Hybrid Force-Position Locomotion Policy (HFPLP) learning framework, where the action space of the policy is defined as a combination of target joint positions and feedforward torques, enabling the robot to rapidly respond to payload variations and external disturbances.
In addition, the proposed Disturbance-Aware Adaptive Compensation (DAAC) provides compensation actions in the torque space based on external disturbance estimation, enhancing the robot's adaptability to dynamic environmental changes.
We validate our approach in both simulation and real-world deployment, demonstrating that it outperforms existing methods in carrying payloads and resisting disturbances.
\end{abstract}


\section{Introduction}\label{Introduction}
Compared to wheeled and tracked robots, legged robots demonstrate advanced adaptability in unstructured environments, enabling more complex locomotion behaviors such as walking\cite{miki2022learning}, jumping\cite{cheng2024extreme}, and climbing\cite{vogel2024robust}.
This remarkable flexibility allows them to navigate narrow and obstacle-dense terrains more effectively, making them highly suitable for applications in rescue missions\cite{kleiner2007real} and space exploration\cite{arm2023scientific}.
In recent years, Reinforcement Learning (RL)-based methods have made significant progress in the field of motion control of legged robots\cite{rudin2022learning,nahrendra2023dreamwaq}.
Unlike traditional control methods that rely on precise modeling\cite{bjelonic2022offline,grandia2023perceptive}, RL can automatically acquire highly dynamic motion skills through extensive simulation training.
However, RL-based motion policies are highly sensitive to payload variations and external disturbances in the real world, posing significant challenges for robots operating in complex environments\cite{peng2018sim,hwangbo2019learning}.
Due to the complexity of working environments and the diversity of tasks, robots inevitably encounter external disturbances during the execution of the task\cite{arm2023scientific,biswal2021development}.
Therefore, enhancing the disturbance rejection capability of motion policies is essential to ensure the efficient and reliable operation of robots in practical applications\cite{lee2024learning,kang2024external}.

External disturbances typically act on the robotic system in the form of forces, leading to changes in system dynamics and affecting its stability\cite{chen2023quadruped}.
However, existing RL-based motion policies for legged robots primarily use target joint positions as the action space\cite{miki2022learning,cheng2024extreme,vogel2024robust}.
The changes in joint positions caused by disturbances can only indirectly reflect the influence of disturbances, but cannot accurately capture the impact of force and torque changes on system stability\cite{soni2023end,queeney2024gram}.
This inherent delay undermines the real-time responsiveness and effectiveness of disturbance rejection\cite{li2024reinforcement}.
In contrast, operating in force space provides a more direct representation of the sources of disturbances and their influence on robot dynamics\cite{chen2023learning}.
Inspired by the classical model-based control framework, which integrates feedforward and feedback mechanisms to compute target joint torques\cite{bjelonic2022offline,grandia2023perceptive,chen2023quadruped}, we propose a novel learning framework.
This framework incorporates feedforward torques and defines the action space of the policy as a combination of the target joint positions and the feedforward torques.
Compared to policies that rely exclusively on joint positions\cite{rudin2022learning,nahrendra2023dreamwaq}, feedforward torques directly account for the impact of robot dynamics and the external environment.
This enables the robot to respond more rapidly to payload variations and external disturbances, thus improving its adaptability to dynamic environmental changes\cite{lyu2024rl2ac}.

In addition, to improve the robustness of motion policies under environmental disturbances, existing methods typically introduce random external forces during training to learn robust policies applicable to various disturbance conditions\cite{hartmann2024deep,shi2024rethinking}.
However, policies trained using this approach lack the ability to dynamically perceive and actively compensate for unknown disturbances\cite{long2024learning}.
Furthermore, for the sake of global optimality, these policies generally exhibit conservatism, resulting in a significant decrease in motion performance in undisturbed environments\cite{kumar2021rma}.
In contrast, an effective solution is to use proprioceptive sensory data to estimate external disturbance information and develop an adaptive compensation policy that actively adjusts the control inputs to compensate for the disturbance\cite{zhu2023proprioceptive}.
This approach has been extensively studied in the model-based control field, demonstrating promising performance\cite{amanzadeh2024predictive}.
However, in practical applications, model-based approaches face several challenges, such as the complexity of constructing non-linear models, high computational cost, and limited scalability\cite{sombolestan2024adaptive}.
Therefore, we propose a learning-based approach that leverages the powerful function approximation capability of Deep Neural Networks (DNNs) to explicitly estimate external disturbance and train an adaptive compensation policy in the joint torque space, enabling real-time perception and active compensation of unknown disturbances.

In this paper, we propose a novel Hybrid Force-Position Locomotion Policy (HFPLP) learning framework to improve the robustness of legged robots under external disturbances.
By incorporating feedforward torques into the action space, HFPLP reduces the reliance on the accuracy of the actuator PD control model\cite{hwangbo2019learning} and significantly improves its dynamic response to payload variations and external impacts.
In addition, we design a Disturbance-Aware Adaptive Compensation (DAAC) mechanism, which explicitly estimates external disturbance and incorporates an active compensation policy in the joint torque space to further optimize the robot's adaptability to unknown disturbances.
Extensive experiments are conducted on the Unitree Go2 quadruped robot\cite{example_website} over various rough terrains with simultaneous external disturbances.
The results demonstrate that our method significantly enhances the robot's robustness to external disturbances in complex environments.

In summary, the contributions of this work are as follows:
\begin{itemize}
\item A novel hybrid force-position locomotion policy learning framework is proposed, which incorporates feedforward torque into the action space to improve the dynamic response of the policy to external disturbances.
\item A disturbance-aware adaptive compensation mechanism is designed, which explicitly estimates the external disturbance and combines it with an active compensation policy in the joint torque space, enhancing the robot's adaptability to unknown disturbances.
\item Extensive experimental results on the real robot demonstrate that our method significantly enhances the robot's payload capacity and improves its robustness against external disturbances in complex environments.
\end{itemize}

\begin{figure*}[t]
\centerline{\includegraphics[width=17cm]{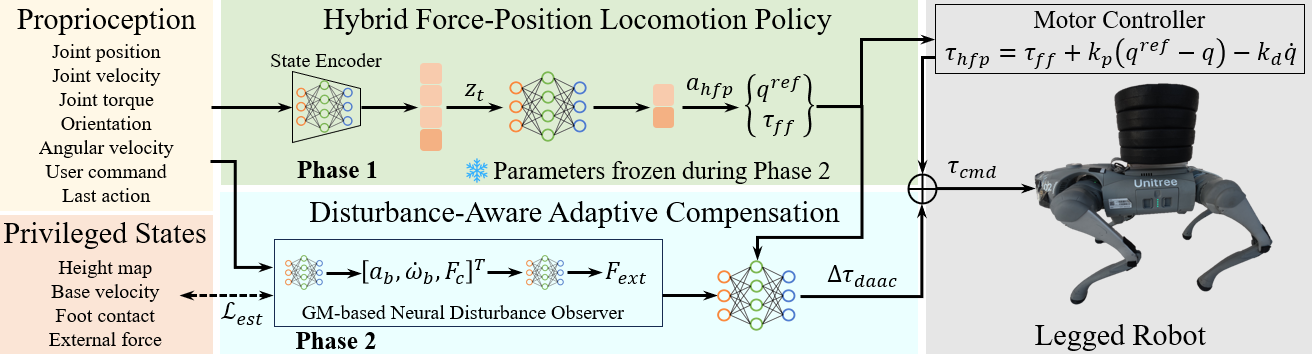}}
\caption{Overview of the training framework. We first employ an asymmetric actor-critic architecture to train the HFPLP, enabling it to output target joint positions and feedforward torques based on proprioception and user commands. These actions are then converted into target joint torques via the motor controller to achieve robust locomotion. Subsequently, we freeze the parameters of the well-trained HFPLP and train the DAAC with a GM-based neural disturbance observer from scratch. The DAAC is designed to generate compensation torques in the joint torque space based on disturbance information, enhancing the robot's adaptability to external disturbances.}
\label{fig1}
\end{figure*}

\section{Related Work}\label{Related}
\subsection{Action Space of RL-based Locomotion Policy}\label{Related-1}
In recent years, methods that use target joint positions as the action space have been widely applied in the motion learning of legged robots and have demonstrated outstanding performance in various tasks\cite{cheng2024extreme,vogel2024robust,rudin2022learning,nahrendra2023dreamwaq}.
However, this paradigm is highly dependent on the accuracy of the actuator model\cite{hwangbo2019learning}.
Differences in PD controller parameter settings and the mismatch between actuator models in simulation and reality significantly impact the performance of locomotion policy on real robots\cite{kumar2021rma,xie2021dynamics}.
On the other hand, researchers have explored policies that utilize target joint torques as the action space\cite{chen2023learning,kim2023torque}.
These policies directly output joint torques for control, avoiding reliance on PD controllers.
This approach allows for a better capture of the dynamic characteristics of the robot and enhances its robustness to external disturbances\cite{sood2023decap}.
However, the learning process in the torque space typically suffers from low sample efficiency and does not consistently converge to natural gaits, which limits the practical application of this method and has prevented its widespread adoption\cite{esseraction}.
In addition, some methods define the action space of the policy as adjusting the parameters of the model-based controller, combining RL with model-based controller to adapt to complex environments and tasks\cite{gangapurwala2022rloc}.
Bellegarda et al.\cite{bellegarda2022cpg} develop a policy to regulate the parameters of Central Pattern Generators (CPGs) to generate robust quadrupedal locomotion.
Miki et al.\cite{miki2022learning} encode temporal components in the action space to guide periodic gaits.
The literature\cite{chen2024learning} introduces an RL-augmented model predictive control (MPC) framework, where the agent learns acceleration compensation to enhance the performance of the MPC.
Unlike the above methods, we propose a novel HFPLP that combines the efficiency of position control in trajectory generation with the flexibility of force control in disturbance compensation\cite{lyu2024rl2ac, bledt2018cheetah}.

\subsection{Robust Locomotion Controller for Legged Robots}\label{Related-2}
Legged robot motion controllers exhibit vulnerabilities when responding to payload variations and unpredictable environmental disturbances, significantly limiting their practical application in complex scenarios\cite{kang2024external}.
To address this issue, researchers have proposed various methods to improve the robustness of motion controllers, aiming to improve the locomotion performance of legged robots in complex environments\cite{lyu2024rl2ac}.
Classical model-based adaptive control methods achieve adaptability to system variations and uncertainties by estimating uncertain parameters in the robot's dynamic model online and designing adaptive laws to adjust the controller parameters accordingly\cite{amanzadeh2024predictive,sombolestan2024adaptive,zhu2023proprioceptive}.
Furthermore, Chen et al.\cite{chen2023quadruped} propose a framework based on capturability analysis to synthesize push recovery controllers, enhancing the robot recovery performance in response to impact disturbances during dynamic motion.
However, these model-based approaches encounter high computational burdens and potential stability issues in practical applications\cite{lee2024learning}.
In recent years, RL-based robust policies have attracted significant attention\cite{shi2024rethinking,long2024learning}.
Leveraging large-scale training data and simulation environments, these approaches aim to learn optimal control policies for robots under varying disturbance conditions\cite{xiao2024pa}. 
Hartmann et al.\cite{hartmann2024deep} incorporate weights for disturbance recovery capabilities into the reward function, successfully training control policies capable of adapting to external disturbances. 
Moreover, Robust Adversarial Reinforcement Learning (RARL) has emerged as an effective method to improve policy robustness, improving disturbance rejection performance of legged robots on complex terrains\cite{shi2024rethinking,long2024learning}.
However, the design of disturbance-related rewards relies heavily on expert knowledge.
This paper implements explicit estimation and adaptive compensation of external disturbances in the force space, further improving the robot's adaptability to unknown disturbances.
This approach offers new perspectives and possibilities for the application of legged robots in complex environments.

\section{Preliminaries}\label{Preliminaries}
\subsection{Reinforcement Learning}\label{Preliminaries-A}
Learning-based motion control problems are typically formulated as a Markov Decision Process (MDP), which enables robots to learn optimal motion policies through interaction with the environment.
The MDP is defined by a tuple $(\mathcal{S},\mathcal{A},\mathcal{P},R,\gamma)$, where $\mathcal{S}$ represents the state space, $\mathcal{A}$ is the action space, $\mathcal{P}(s^{'}|s,a)$ is the state transition probability, $R(s,a)$ denotes the reward function, and $\gamma\in[0,1)$ is the discount factor.
The robot’s state $s\in \mathcal{S}$ encodes proprioceptive information, while the action $a\in \mathcal{A}$ corresponds to the control input.
The objective of the RL agent is to learn an optimal policy $\pi^{*}(a|s)$ that maximizes the expected cumulative reward:
\begin{equation}\label{eq1}
J(\pi)=\mathbb{E}_{\pi}\left[\sum_{t=0}^{\infty}\gamma^{t}R(s_{t},a_{t})\right],
\end{equation}
where the reward function is designed to encourage desired behaviors, effectively guiding the learning process towards achieving task-specific objectives.

\subsection{Dynamics of Legged Robots}\label{Preliminaries-B}
Let $q$ represents the generalized coordinates, the basic dynamics of the robot can be written as:
\begin{equation}\label{eq2}
\bm{M}\bm{\ddot{q}}+\bm{C}\bm{\dot{q}}+\bm{G}=\bm{S}^{T}\bm{\tau}+\bm{J}_{c}^{T}\bm{F}_{c},
\end{equation}
where $\bm{M},\bm{C\dot{q}},\bm{G},\bm{S}^{T},\bm{\tau},\bm{J}_{c}^{T},\bm{F}_{c}$ are the inertia matrix, Coriolis force, gravity force, selection matrix, joint torque input, contact Jacobian matrix, and contact force, respectively.
When the payload varies or external forces are applied, the system dynamics deviate from the basic model.
To maintain system stability, it is typically necessary to apply additional torques to compensate for the effects of unmodeled external disturbances.
The corresponding system dynamics can be represented as:
\begin{equation}\label{eq3}
\bm{M\ddot{q}}+\bm{C\dot{q}}+\bm{G}=\bm{S}^{T}(\bm{\tau}+\Delta\bm{\tau})+\bm{J}_{c}^{T}\bm{F}_{c}+\bm{J}_{ext}^{T}\bm{F}_{ext},
\end{equation}
where $\Delta\bm{\tau}$ represents the compensation torque, $\bm{F}_{ext}$ and $\bm{J}_{ext}$ represent the external disturbances acting on the robot’s body and the corresponding Jacobian matrix.
In the RL-based control framework, the policy $\pi$ receives the state $s$ as input and outputs an action $a$, which is then transformed by the actuators into joint torques $\tau$ to accomplish the specific task.

\section{Methods}\label{Methods}
This section provides a detailed description of the construction and training of HFPLP and DAAC.
As illustrated in Fig.~\ref{fig1}, the framework employs a two-stage training approach.
In the first stage, the HFPLP is trained using an asymmetric actor-critic structure, enabling the robot to simultaneously adjust foot position trajectories and generate appropriate feedforward forces.
In the second stage, the well-trained HFPLP parameters are frozen and integrated as part of the environment dynamics.
Subsequently, the DAAC is trained from scratch.
This policy employs a Generalized Momentum(GM)-based neural disturbance observer to perceive external disturbances and outputs compensation torques in the joint torque space, improving the robot's adaptability to external disturbances.

\subsection{Hybrid Force-Position Locomotion Policy}\label{Methods-A}
Legged robots are complex underactuated systems that precisely control the dynamic interaction between the foot and the environment by applying appropriate joint torques through joint actuators, achieving stable locomotion tasks.
Classic model-based control methods integrate trajectory planning with robot dynamics to achieve swing leg trajectory tracking and stance leg support force optimization\cite{bledt2018cheetah}.
The robot's joint actuator model can be expressed as:
\begin{equation}\label{eq4}
\bm{\tau}_{i}=\bm{\tau}_{i,ff}+\bm{J}_{i}^{T}[\bm{K}_{p}(\bm{p}_{i}^{ref}-\bm{p}_{i})+\bm{K}_{d}(\bm{v}_{i}^{ref}-\bm{v}_{i})],
\end{equation}
where $\bm{\tau}_{i,ff}$ represents the feedforward torque, $\bm{J}_{i}$ denotes the foot Jacobian, $\bm{K}_{p}$ and $\bm{K}_{d}$ are proportional and derivative gains, $\bm{p}_{i}^{ref}$ and $\bm{v}_{i}^{ref}$ indicate the reference foot position and velocity, $\bm{p}_{i}$ and $\bm{v}_{i}$ represent the actual foot position and velocity, $i\in\{1,\cdots,n_{l}\}$, $n_{l}$ denotes the number of legs.
Typically, stance legs are more concerned with feedforward torques to counteract gravity and external disturbances, while swing legs prioritize PD controller tracking of the foot position.

However, as discussed in Section \ref{Related-1}, existing RL-based motion policies focus primarily on joint positions or joint torques.
These policies couple foot position tracking and foot force control within a single action space, which limits the system's adaptability to environmental disturbances and results in poor interpretability of the actions output by motion policies\cite{lyu2024rl2ac}.
In this work, we propose a novel HFPLP that combines the advantages of swing leg trajectory tracking and stance leg support force optimization.
In order to implement this objective, the action space for HFPLP policy is formulated as:
\begin{equation}\label{eq5}
\bm{a}_{hfp}=[\bm{q}^{ref},\bm{\tau}_{ff}]^{T},
\end{equation}
where $\bm{q}^{ref}$ represents the target joint position and $\bm{\tau}_{ff}$ represents the feedforward torque.
The joint actuator model corresponding to this action space is:
\begin{equation}\label{eq6}
\bm{\tau}_{hfp}=\bm{\tau}_{ff}+\bm{K}_{p}(\bm{q}^{ref}-\bm{q})-\bm{K}_{d}\bm{\dot{q}},
\end{equation}
where $\bm{\tau}_{hfp}$ represents the joint torque generated by HFPLP.

\subsection{Disturbance-Aware Adaptive Compensation}\label{Methods-B}
In complex environments, legged robots may encounter external disturbances, such as unexpected foot contact events or payload variations.
External disturbances cause dynamics shift of the robot system, leading the actual motion trajectory to deviate from the desired trajectory.\cite{lyu2024rl2ac}.
In model-based control, the GM-based disturbance observer is widely employed to estimate external contact forces.
In Eq.~\eqref{eq3}, the sum of all external forces is represented as the disturbance vector:
\begin{equation}\label{eq7}
\bm{\tau}_{d}=\bm{J}_{c}^{T}\bm{F}_{c}+\bm{J}_{ext}^{T}\bm{F}_{ext}.
\end{equation}
As described in\cite{bledt2018contact}, the disturbance estimation $\hat{\bm{\tau}}_{d}$ after passing through a discrete-time low-pass filter can be calculated as follows:
\begin{equation}\label{eq8}
\begin{split}
\hat{\bm{\tau}}_{d}&=\beta\bm{p}-\frac{1-\gamma}{1-\gamma z^{-1}}(\beta\bm{p}+\bm{S}^{T}\bm{\tau}+\bm{C}^{T}\dot{\bm{q}}-\bm{G}), \\
\bm{p}&=\bm{M}\dot{\bm{q}},\beta=\frac{(1-\gamma)\gamma^{-1}}{\Delta t},\gamma=e^{-\lambda\Delta t},
\end{split}
\end{equation}
where $z$ is the z-domain variable, $\Delta t$ represents the sampling period, and $\lambda$ represents the cutoff frequency of the filter.
This method avoids the direct measurement of acceleration $\ddot{\bm{q}}$ by employing summation by parts and generalized momentum $\bm{p}$.

In this work, we propose a GM-based neural disturbance observer that can estimate unknown external force disturbances in real time without relying on an accurate model.
External disturbances are modeled as forces acting on the robot's Center of Mass (CoM), while torques arising from the offset of the disturbance application point are neglected.
Initially, the proprioceptive history is used to estimate the acceleration vector $[\bm{a}_{b}, \bm{\dot{\omega}}_{b}]$ and the foot contact force $\bm{F}_{c}$.
The acceleration vector is calculated by differencing the velocity in the simulation.
Subsequently, these estimates, along with current proprioceptive observation, are fed into the disturbance observer to estimate the external disturbance force $\bm{F}_{ext}$.

In addition, we develop an adaptive compensation policy to mitigate the impact of disturbance forces on the system.
Benefiting from the HFPLP introduced in Section \ref{Methods-A}, which exhibits sensitivity to feedforward forces, the adaptive policy can directly perform disturbance compensation in the joint torque space.
As illustrated in Fig.~\ref{fig1}, the parameters of the well-trained HFPLP are frozen, and the DAAC is learned from scratch to output the compensation torque $\Delta \bm{\tau}_{daac}$. 
This compensation torque is then combined with the torque output from the HFPLP to serve as the joint torque command for the robot:
\begin{equation}\label{eq9}
\bm{\tau}_{cmd}=\bm{\tau}_{hfp}+\Delta \bm{\tau}_{daac}.
\end{equation}

\subsection{Training Details}\label{Methods-C}
All policies are trained using the Proximal Policy Optimization (PPO) algorithm in the Isaac Gym simulation\cite{rudin2022learning} and zero-shot transferred to the Unitree Go2 quadruped robot.
In the simulation, the magnitude and duration of the disturbance force are randomly sampled to mimic external forces in the real world.
The $3$-dim disturbance is sampled in $([-100,100]\times[-100,100]\times[-200,0])\,N$, and the duration time is sampled from $(1,4)\,s$.
Additionally, in each episode, $60\,\%$ of the environments are randomly selected to experience disturbance forces.

\subsubsection{Learning the HFPLP}\label{Methods-C-1}
The proposed HFPLP uses proprioceptive information and base velocity command as inputs, and outputs target joint positions and feedforward torques to achieve the desired velocity tracking for the legged robot.
The proprioceptive observation $\bm{o}_{t}\in\mathbb{R}^{69}$ contains body angular velocity $\bm{\omega}_{t}$, projected gravity vector $\bm{g}_{t}$, base velocity command $\bm{c}_{t}$, joint positions $\bm{q}_{t}$, joint velocities $\dot{\bm{q}}_{t}$, joint torques $\bm{\tau}_{t}$, and previous action $\bm{a}_{t-1,hfp}$.
The history length $H=5$.
The critic network incorporates additional privileged information, including base velocity $\bm{v}_{b}$, height map $\bm{h}_{t}$, foot contact $\bm{c}_{f}$, foot clearance $\bm{p}_{f}$, and disturbance forces $\bm{F}_{ext}$.
The action $\bm{a}_{t,hfp}\in\mathbb{R}^{24}$ includes target joint position $\bm{q}^{ref}$ and feedforward torque $\bm{\tau}_{ff}$, which are transformed into the joint torque by Eq.~\eqref{eq6} with $K_{p}=20$ and $K_{d}=0.5$.
The reward functions follow the work of DreamWaQ\cite{nahrendra2023dreamwaq}, where a new reward function $(\bm{q}^{ref}-\bm{q})^{2}$ is added to encourage joint position tracking with a weight of $-0.2$.
Curriculum learning\cite{zhang2024robust} is used to gradually increase the difficulty of tasks, allowing the robot to adapt to rough terrain.

\subsubsection{Learning the DAAC}\label{Methods-C-2}
The disturbance observer is composed of two concatenated $3$-layer MLP with ReLU activations.
The first MLP takes in historical observations $\bm{o}_{t}^{H}\in\mathbb{R}^{345}$ as input and outputs $[\bm{a}_{b}, \bm{\dot{\omega}}_{b},\bm{F}_{c}]^{T}\in\mathbb{R}^{18}$, with hidden layer $[256\;128\;64]$. 
The second MLP receives $[\bm{o}_{t},\bm{a}_{b}, \bm{\dot{\omega}}_{b},\bm{F}_{c}]^{T}\in\mathbb{R}^{87}$ as input and outputs $\bm{F}_{ext}\in\mathbb{R}^{3}$, with hidden layer $[256\;128\;64]$.
The observation of the DAAC $o_{t,daac}\in\mathbb{R}^{96}$ includes proprioceptive observation $\bm{o}_{t}$, disturbance force $\bm{F}_{ext}$, and action $\bm{a}_{t,hpf}$.
The action $\bm{a}_{t,daac}\in\mathbb{R}^{12}$ represents the compensation torque $\Delta \bm{\tau}_{daac}$, which is transformed into the joint torque command by Eq.~\eqref{eq9}. 
The reward functions are consistent with HFPLP training.

\section{Experiments}
To evaluate the locomotion performance and disturbance adaptation of the proposed method, the state-of-the-art RL-based locomotion method DreamWaQ\cite{nahrendra2023dreamwaq} is utilized as the baseline.
Comparative experiments and ablation studies are conducted to demonstrate the improvements over the baseline.
The following are the baseline methods in the experiment, including ablation studies to demonstrate the necessity of some submodule of our method:
\begin{itemize}
\item \textbf{DreamWaQ-FT}: Fine-tuning the DreamWaQ in environments with random disturbances.
\item \textbf{DreamWaQ-DAAC}: Learning the DAAC based on the DreamWaQ.
\item \textbf{HFPLP-FT}: Fine-tuning the HFPLP in environments with random disturbances.
\item \textbf{HFPLP-DAAC w/o DO}: Learning the DAAC without disturbance observer based on the HFPLP.
\item \textbf{HFPLP-DAAC (Ours)}: Learning the DAAC based on the HFPLP.
\end{itemize}

\subsection{Simulation Results}\label{Experiments-A}
\subsubsection{Action Tracking}
\begin{figure}[t]
\centerline{\includegraphics[width=8.5cm]{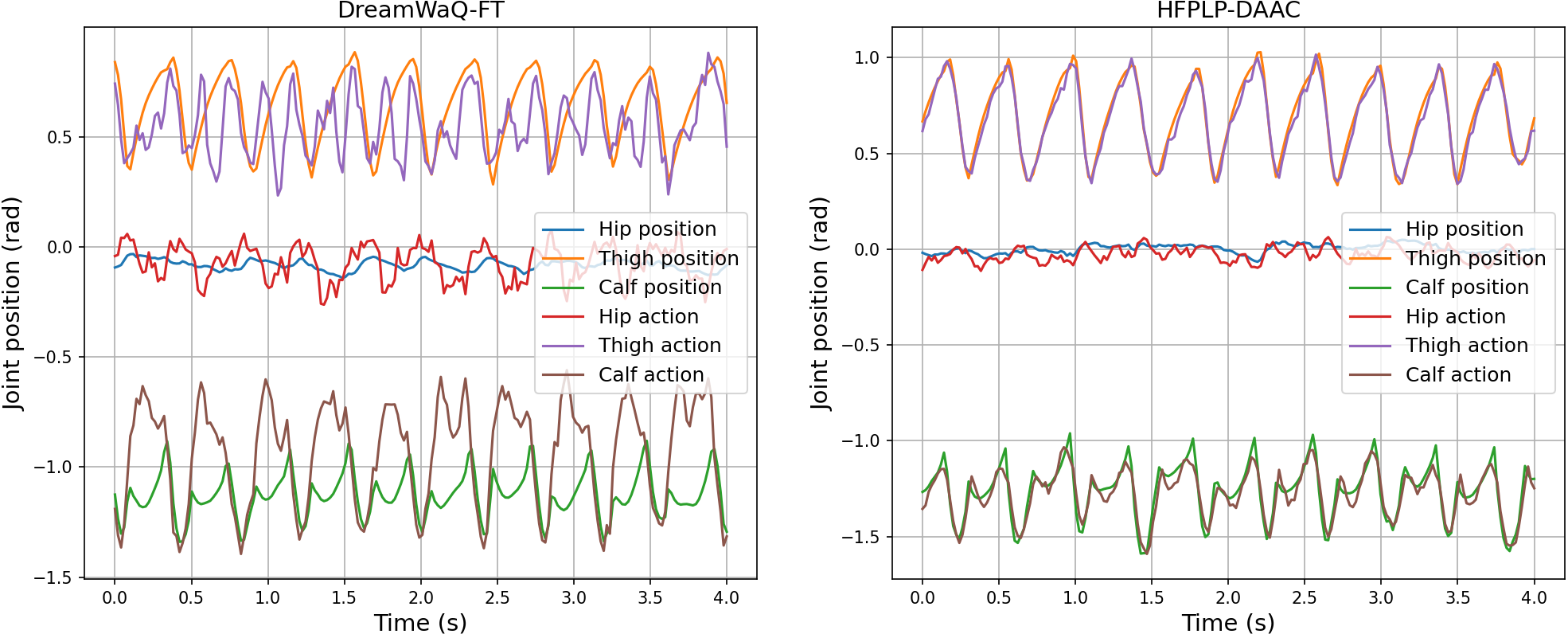}}
\caption{The joint position tracking of the front right leg during robot motion at $v_{x}=1\,m/s$. Due to the additional contact forces required by the stance leg to provide gravity compensation, the baseline exhibits significant joint position tracking errors. In contrast, our proposed method significantly improves the tracking accuracy by providing feedforward torques.}
\label{exp1}
\end{figure}

To demonstrate the advantages of incorporating feedforward torque in HFPLP, we compare the joint position tracking curves of DreamWaQ-FT and HFPLP-DAAC.
As shown in Fig.~\ref{exp1}, DreamWaQ-FT compensates for contact forces by adjusting joint position tracking errors of the stance leg.
In contrast, HFPLP-DAAC incorporates feedforward torque into its actions, significantly improving joint position tracking accuracy, demonstrating that HFPLP combines the advantages of swing leg position control and stance leg force control to effectively provide compensation torque in the stance phase.

\subsubsection{Disturbance Adaptation}
\begin{figure}[t]
\centerline{\includegraphics[width=8.5cm]{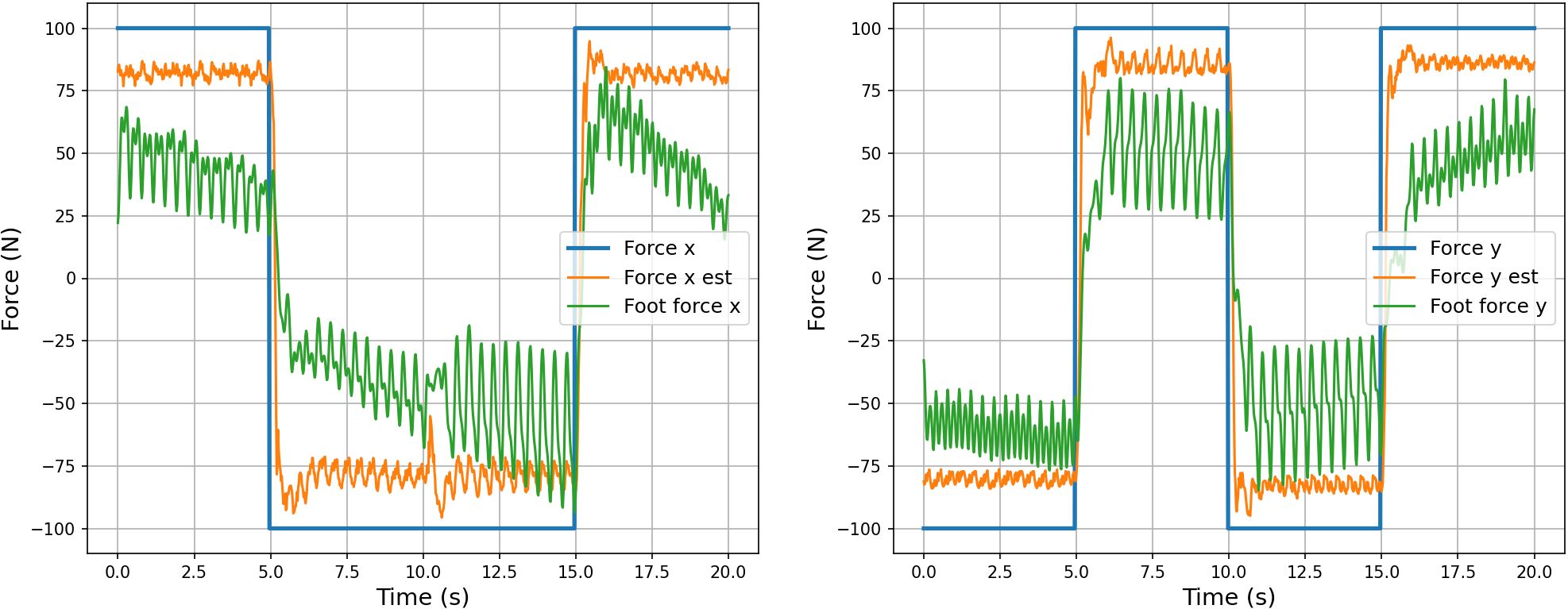}}
\caption{Adaptation of the DAAC to disturbances. The blue line represents the applied disturbance force, the orange line represents the estimated force from the disturbance observer, and the green line represents the foot force generated by the compensation torque of the DAAC through the foot Jacobian mapping. The DAAC effectively generates compensation torque based on the disturbance force estimated by the disturbance observer.}
\label{exp2}
\end{figure}

\begin{figure}[t]
\centerline{\includegraphics[width=8.5cm]{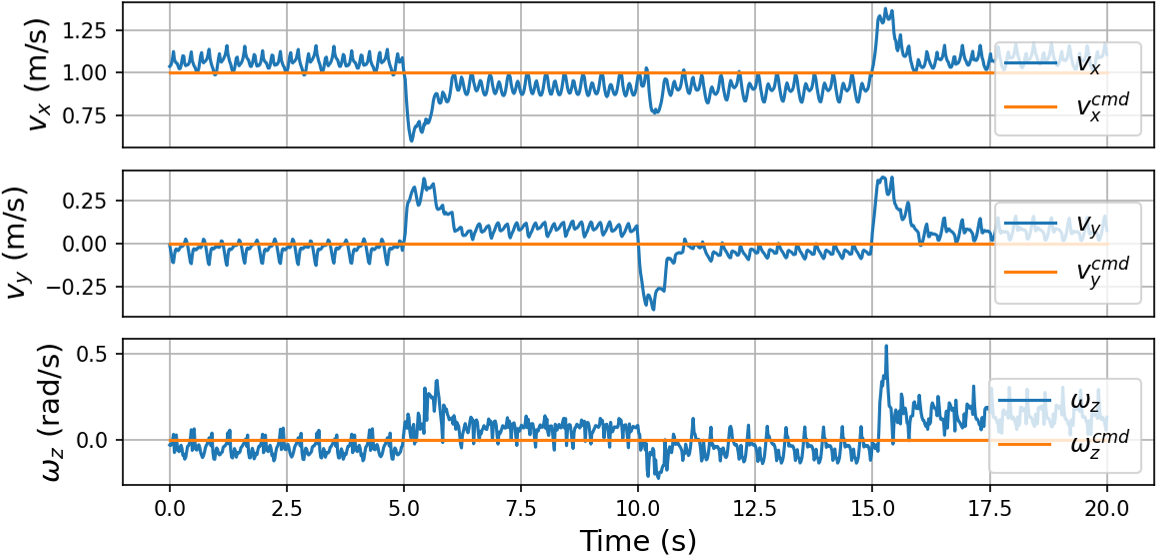}}
\caption{Velocity tracking curve under external disturbances. The HFPLP-DAAC demonstrates the ability to rapidly recover from external disturbances and adapt to the disturbance force to maintain the desired motion state.}
\label{exp3}
\end{figure}

In the simulation, time-varying disturbance forces are introduced to evaluate the disturbance adaptation of HFPLP-DAAC.
In order to analyze the response of DAAC to disturbance, the compensation torques are mapped to foot forces using the foot Jacobian: $\bm{F}_{ee}=\bm{J}^{-T}\Delta \bm{\tau}$.
The disturbance forces along the $X$ and $Y$ axes are applied with magnitudes of $100\,N$, and their directions are changed at intervals of $5\,s$.
Fig.~\ref{exp2} illustrates the applied disturbance forces, the disturbance forces estimated by the disturbance observer, and the compensation foot forces generated by the DAAC.
It can be observed that the disturbance observer is able to predict disturbance forces, and compensation foot forces generated by the DAAC align with the trends of the disturbance forces.
Consequently, the corresponding ground reaction forces effectively counteract the impact of the disturbance forces.
This is attributed to the compensation mechanism introduced in the force space, which enhances the interpretability and transparency of the DAAC.
Additionally, the velocity tracking performance of the robot under disturbance forces is shown in Fig.~\ref{exp3}, demonstrating that the HFPLP-DAAC can enable the robot to rapidly recover to the desired motion state when subjected to significant disturbances.

\subsection{Ablation Study}\label{Experiments-B}
\begin{figure}[t]
\centerline{\includegraphics[width=8.5cm]{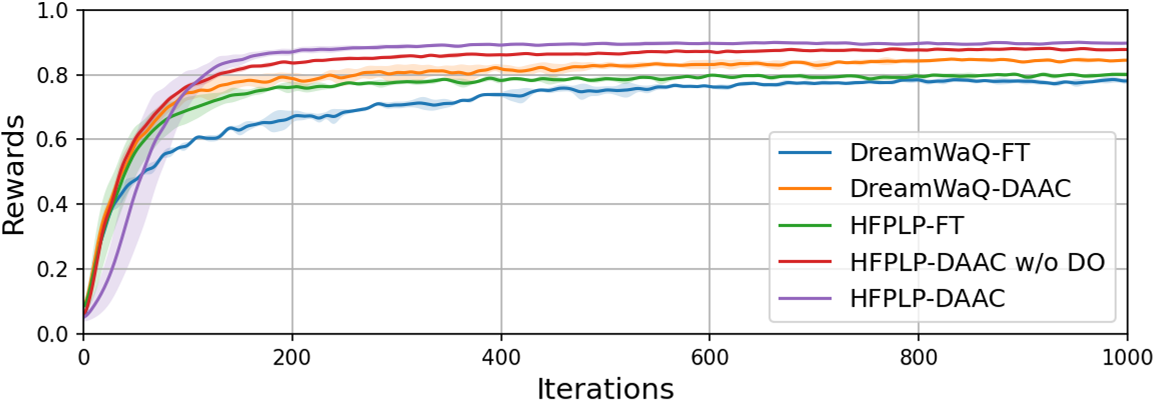}}
\caption{Learning curves of different methods. The curves and shaded areas represent the mean and standard deviation of rewards across $10$ different seeds, respectively. The HFPLP-DAAC outperforms baseline and other ablation versions on rewards, indicating higher motion performance.}
\label{exp4}
\end{figure}

To evaluate the impact of different modules on the robot's motion performance, ablation experiments are conducted.
All policies are trained under identical environmental setups and hyper-parameters to ensure fairness in the results.
Fig.~\ref{exp4} illustrates the learning curves of different policies, demonstrating that HFPLP-DAAC outperforms the baseline and other ablation versions.
The comparison between DreamWaQ-FT and HFPLP-FT shows that the introduction of feedforward action space improves the response of the policy to disturbances.
The comparison between HFPLP-FT and HFPLP-DAAC highlights that the DAAC leverages estimated disturbance information to output additional compensation torques, thereby enhancing the overall performance.
The performance degradation observed in the absence of the disturbance observer underscores the importance of explicit disturbance estimation in improving disturbance responsiveness.

\begin{figure}[t]
\centerline{\includegraphics[width=8.5cm]{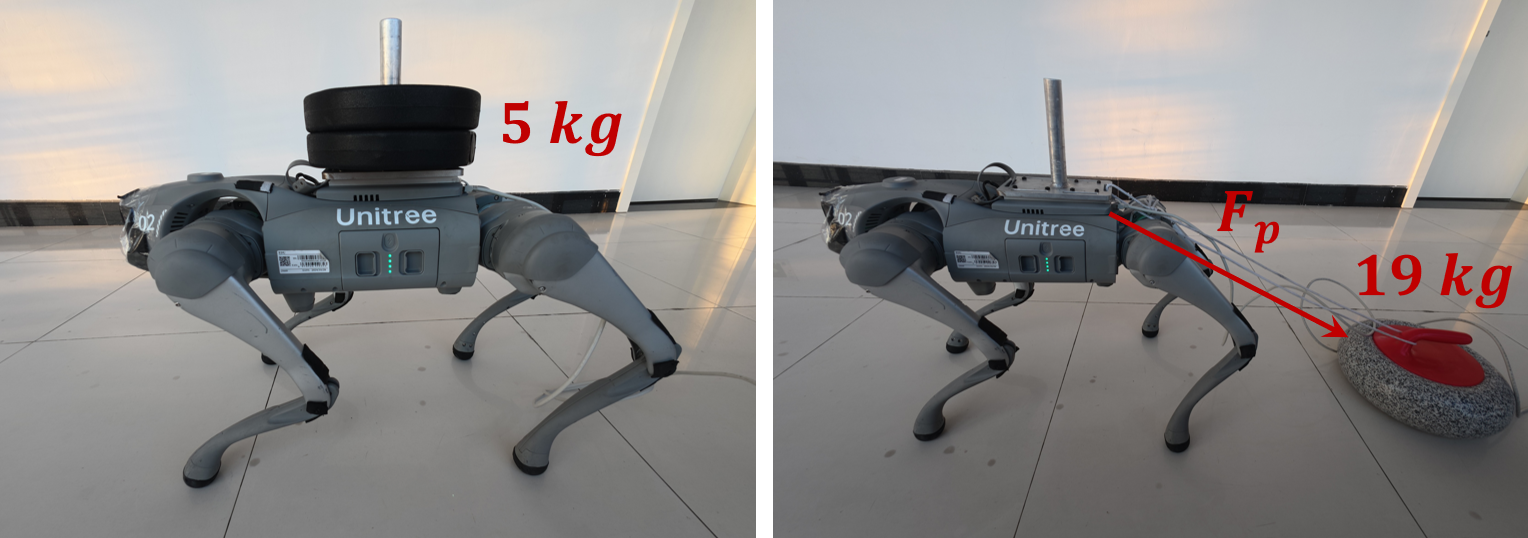}}
\caption{Diagram of the real world ablation experiment setup. Left: The robot carries a $5\,kg$ payload. Right: The robot pulls a $19\,kg$ weight via a rope to simulate pulling disturbance.}
\label{exp5}
\end{figure}

\begin{figure}[t]
\centerline{\includegraphics[width=8.5cm]{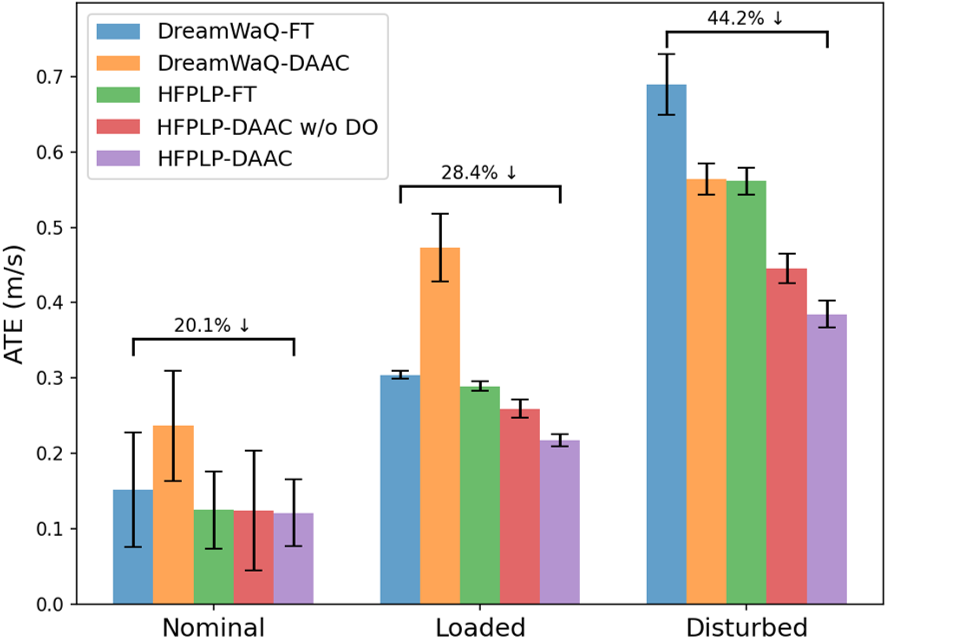}}
\caption{Comparison of ATE under different disturbance conditions for various policies. The HFPLP-DAAC significantly improves velocity tracking performance under load and external force disturbances. Moreover, even in the absence of disturbances, our method outperforms the baseline, demonstrating superior sim-to-real performance.}
\label{exp6}
\end{figure}

In addition, all policies are deployed on a real robot to evaluate their performance in the real world.
For each policy, the Absolute Tracking Error (ATE) is calculate under three conditions: nominal condition, loaded condition, and disturbed condition with pulling disturbance.
The robot follows $v_{x}=1\,m/s$, $v_{y}=1\,m/s$ commands sequentially, and each task is executed $5$ times.
The experimental setups for payload and lateral pulling disturbance are shown in Fig.~\ref{exp5}, where the payload is $5\,kg$ and the weight dragged by the rope is $19\,kg$.
The average pulling force $F_{p}$ exerted by the rope is approximately $40\,N$, measured using the digital dynamometer.
Fig.~\ref{exp6} illustrates the ATE under various environmental conditions.
The experimental results demonstrate that the proposed method exhibits superior disturbance rejection capabilities compared to other policies.

\subsection{Disturbance Adaptation on Real Robot}\label{Experiments-C}
\subsubsection{Robustness to Actuators}
\begin{figure}[t]
\centerline{\includegraphics[width=8.5cm]{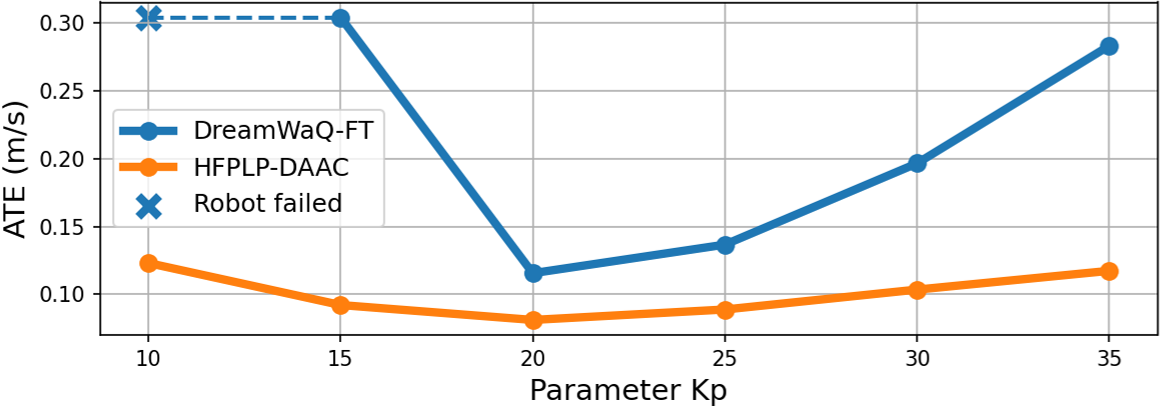}}
\caption{The locomotion performance of the policy trained in simulation with $K_{p}=20$, $K_{d}=0.5$ when transferred to the real robot with different PD parameters. Compared to the baseline, HFPLP-DAAC demonstrates adaptability to different $K_{p}$ values, reducing sensitivity to actuator discrepancies between simulation and real robot.}
\label{exp7}
\end{figure}

The baseline method relies on the actuator's PD controller to convert the target joint position into the desired torques, and differences between the PD controller in simulation and the real robot will lead to sim-to-real problem. 
In contrast, our method reduces dependence on the PD controller through feedforward torque control, which better accommodates the actuator discrepancies between simulation and the real robot, improving motion robustness.
Fig.~\ref{exp7} shows the motion performance of policies trained in simulation with $K_{p}=20$, $K_{d}=0.5$ transferred to the real robot with different PD parameters.
As illustrated in the Fig.~\ref{exp7}, our method can still maintain motion performance under large differences in parameter $K_{p}$, while the performance of the baseline method drops significantly. 
This indicates that our method can avoid tedious PD parameter tuning and reduce the sim-to-real gap.

\subsubsection{Heavy Payload Adaptation}
\begin{figure}[t]
\centerline{\includegraphics[width=8.5cm]{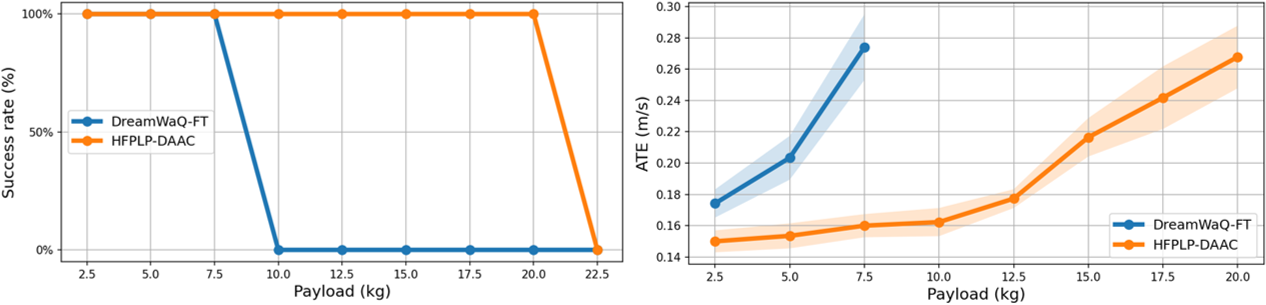}}
\caption{The success rate and ATE of the robot when carrying different payloads. The HFPLP-DAAC demonstrates the ability to successfully track a velocity command of $v_{x}=1.0\,m/s$ while carrying a payload of $20.0\,kg$ (equivalent to $133\,\%$ of its own weight), significantly enhancing the robot's capability to perform tasks with high payloads.}
\label{exp8}
\end{figure}

\begin{figure}[t]
\centerline{\includegraphics[width=8.5cm]{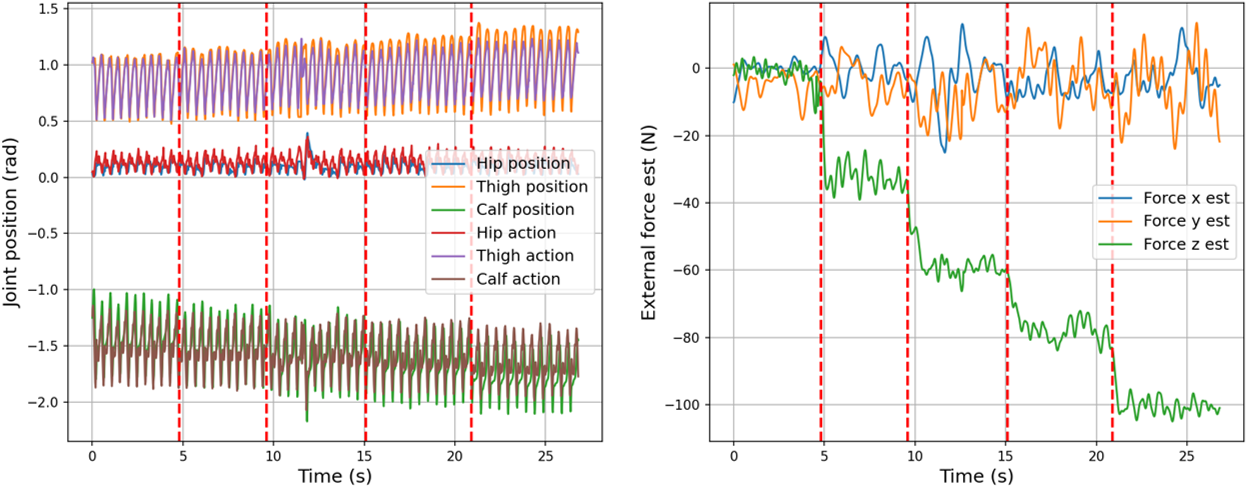}}
\caption{Joint position tracking and external force estimation under dynamically varying payloads. The joint position tracking remains consistent despite changes in the payload, while the disturbance observer accurately estimates the external force disturbances. These results demonstrate that the HFPLP-DAAC effectively perceives variations in external forces and provides appropriate feedforward compensation torques to mitigate the impact of disturbances.}
\label{exp9}
\end{figure}

To demonstrate the adaptability of the motion policy to varying payloads, we evaluate the robot's locomotion performance under different payloads.
The robot tracks a velocity command of $v_{x}=1.0\,m/s$ for a duration of $10\,s$ while carrying various payloads.
For each payload, $5$ trials are conducted to calculate the Success Rate (SR) and ATE.
As illustrated in Fig.~\ref{exp8}, the baseline could only handle a maximum payload of $7.5\,kg$, whereas our proposed method achieve remarkable performance under a payload of $20.0\,kg$ ($133\,\%$ uncertainty).
To the best of our knowledge, this is the first demonstration of a learning-based approach enabling the real robot to handle a payload exceeding its own weight, highlighting the state-of-the-art capability of our method in payload handling.
Additionally, we evaluate the joint position tracking and external force estimation under dynamically varying payloads.
The robot is commanded to track a velocity of $v_{x}=1.0\,m/s$ while the payloads are gradually increased to $[2.5, 5.0, 7.5, 10.0]\,kg$.
As shown in Fig.~\ref{exp9}, during dynamic payload changes, joint position tracking exhibited no significant variation, and the disturbance observer effectively estimates the external force magnitude.
These results demonstrate that the DAAC can accurately perceive changes in external forces and provide effective feedforward compensation torques to maintain the robot's locomotion performance.

\subsubsection{Impact Disturbance Adaptation}
We further evaluate the adaptability of HFPLP-DAAC to impact disturbances.
To ensure a controlled and quantifiable comparison, we introduce lateral impact disturbances by releasing a weight from a fixed height, allowing it to swing laterally and collide with the robot moving at a velocity of $v_{x}=1.0\,m/s$.
Each set of experiments is conducted with $10$ trials. 
We evaluate the performance of the algorithm using three metrics: SR, ATE, and Lateral Displacement (LD).
Table \ref{table1} shows that our method significantly improves the ability to resist impact disturbances.
The robot has smaller ATE and LD under larger impact disturbances, indicating that our method improves the response to impact disturbances and enables rapid recovery.

\begin{table}[H]
\centering
\caption{Comparison of Locomotion Performance Under Impact Disturbances}
\label{table1}
\begin{tabular*}{0.47\textwidth}{@{}ccccc@{}}
\toprule
\multirow{2}{*}{Mehtods} & \multirow{2}{*}{Weights} & \multicolumn{3}{c}{Metrics} \\
\cline{3-5}
 &   & SR ($\%$) & ATE ($m/s$) & LD ($m$)\\
\midrule
\multirow{4}{*}{DreamWaQ-FT} & $2.5\,kg$ & $\mathbf{100}$ & $0.54\pm0.14$ & $0.39\pm0.09$ \\
 & $5.0\,kg$ & $40$ & $0.76\pm0.17$ & $0.94\pm0.16$\\
  & $7.5\,kg$ & $0$ & - & -\\
   & $10.0\,kg$ & $0$ & - & - \\
\hline
\multirow{4}{*}{HFPLP-DAAC} & $2.5\,kg$ & $\mathbf{100}$ & $\mathbf{0.23\pm0.06}$ & $\mathbf{0.10\pm0.03}$\\
 & $5.0\,kg$ & $\mathbf{100}$ & $\mathbf{0.35\pm0.10}$ & $\mathbf{0.23\pm0.04}$\\
 & $7.5\,kg$ & $\mathbf{100}$ & $\mathbf{0.47\pm0.09}$ & $\mathbf{0.41\pm0.07}$\\
  & $10.0\,kg$ & $\mathbf{80}$ & $\mathbf{0.56\pm0.12}$ & $\mathbf{0.57\pm0.11}$\\
\bottomrule
\end{tabular*}
\end{table}

\begin{figure}[t]
\centerline{\includegraphics[width=8.5cm]{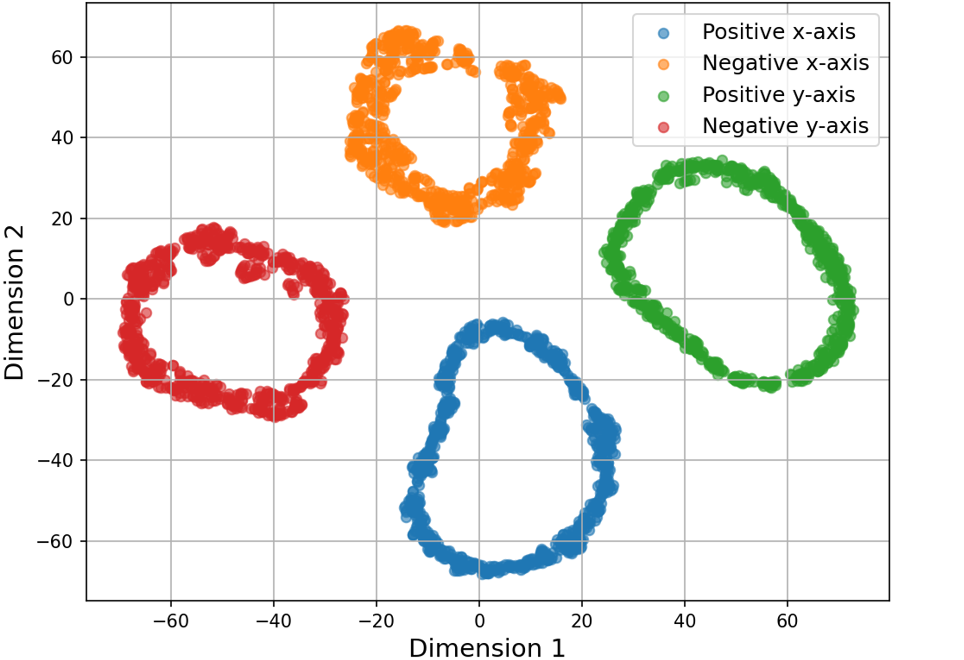}}
\caption{The t-SNE visualization of compensation torques generated by the DAAC. The compensation torques under different directional pulling disturbances are clearly distinguishable, demonstrating that the DAAC effectively perceives the disturbances and generates appropriate compensation torques to mitigate their effects.}
\label{exp10}
\end{figure}

\begin{figure}[t]
\centerline{\includegraphics[width=8.5cm]{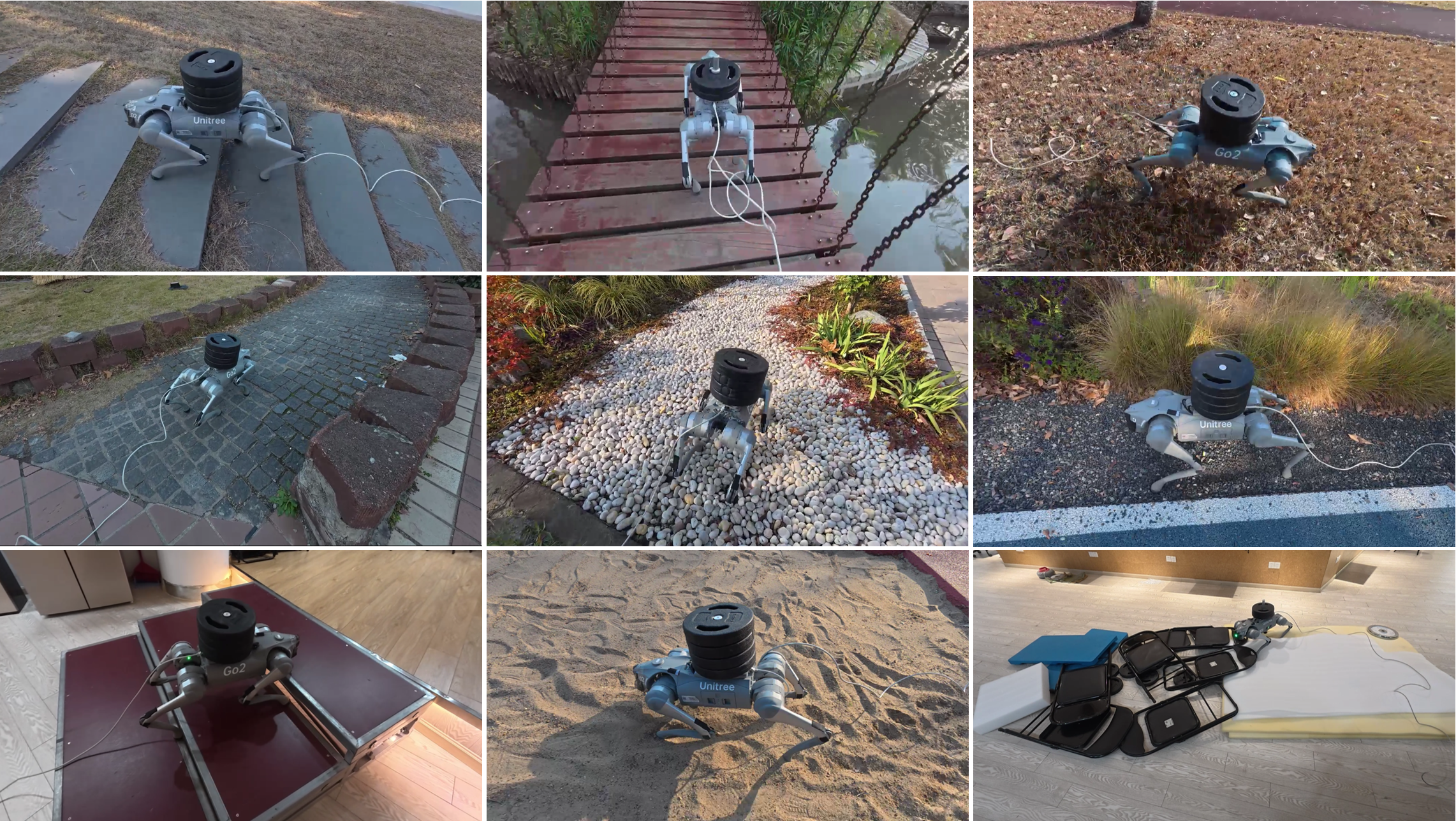}}
\caption{Evaluating the locomotion performance on challenging terrains, including grass, stairs, gravel, soft sand, pebbles, and deformable sponge mats. HFPLP-DAAC significantly improves complex terrain traversal capabilities under payload.}
\label{exp11}
\end{figure}

\subsubsection{Analysis of DAAC Policy}
We employ t-distributed Stochastic Neighbor Embedding (t-SNE) to analyze the actions of the DAAC under various disturbances.
In the experiments, the robot is controlled to move along the positive and negative directions of the x-axis and y-axis, respectively.
Following the method illustrated in Fig.~\ref{exp5}, the robot is subjected to pulling disturbances in different directions.
As shown in Fig.~\ref{exp10}, the actions generated by the DAAC demonstrate a clear distinction between the effects of disturbances from different directions.
This indicates that the DAAC effectively integrates disturbance observer to accurately estimate external forces and provides appropriate feedforward torque compensation, enabling the robot to respond effectively to external disturbances.

\subsubsection{Traversability in Complex Environments}
We demonstrate the locomotion performance of the HFPLP-DAAC under various challenging terrains and external disturbances.
As shown in Fig.~\ref{exp11}, the robot is capable of successfully traversing unstructured terrains such as sandy surfaces, slippery gravel, variable sponge mats, and stair under payload conditions, while effectively handling disturbances such as unexpected collisions and foot slippage.
The experimental results indicate that our approach significantly enhances the robot's ability to traverse complex environments with heavy payloads and improves task execution stability. Further experimental details can be found in the supplementary video.

\section{Conclusion}
In this work, we propose the HFPLP, which significantly improves the dynamic response of legged robots to payload variations and external force impacts.
In addition, benefiting from the sensitivity of HFPLP to the torque space, we introduce the DAAC in the torque space, further enhancing the robot's adaptability to unknown disturbances.
Comprehensive validation using the Unitree Go2 quadruped robot demonstrates that HFPLP-DAAC exhibits superior locomotion performance under various complex terrains and disturbance conditions.
However, this approach still has some limitations, such as the lack of theoretical guarantees and the disturbance model considering only forces applied to the robot's CoM.
Future work will focus on exploring more complex disturbance models and further developing theoretical analysis frameworks to enhance the adaptability of legged robots to various types of disturbances.

\bibliographystyle{IEEEtran}
\balance
\bibliography{main}

\end{document}